%% file: paperX.tex
\documentclass{article}
\usepackage{spconf}
\usepackage{multirow}

\usepackage{graphicx} 
\usepackage{amsmath}
\usepackage{bbm}
\usepackage{ifthen}
\usepackage{adjustbox}
\usepackage{xkeyval}
\usepackage{tikz}
\usepackage{calc}
\usepackage{amssymb}
\usepackage{multirow}
\usepackage{paralist}
\usepackage{comment}
\usepackage{booktabs}       
\usepackage[hidelinks]{hyperref}
\usepackage{multirow}
\usepackage[normalem]{ulem}
\useunder{\uline}{\ul}{}
\usepackage{xspace}
\usepackage{colortbl}
\usepackage{lipsum}

\usepackage[capitalize]{cleveref}
\crefname{section}{Sec.}{Secs.}
\Crefname{section}{Section}{Sections}
\Crefname{table}{Table}{Tables}
\crefname{table}{Tab.}{Tabs.}

\title{Acquisition Time-Informed Breast Tumor Segmentation from Dynamic Contrast-Enhanced MRI}
\name{
Rui Wang$^{1}$,
Yuexi Du$^{1}$,
John Lewin$^{2}$, 
R.~Todd~Constable$^{1,2}$,
Nicha~C.~Dvornek$^{1,2}$
}
\address{
    Departments of 
    $^1$Biomedical Engineering, 
    $^2$Radiology \& Biomedical Imaging, \\
    Yale University, New Haven, CT, USA \\
}
\begin{document}
%
\maketitle
\begin{abstract}

Dynamic contrast-enhanced magnetic resonance imaging (DCE-MRI) plays an important role in breast cancer screening, tumor assessment, and treatment planning and monitoring. The dynamic changes in contrast in different tissues help to highlight the tumor in post-contrast images. However, varying acquisition protocols and individual factors result in large variation in the appearance of tissues, even for images acquired in the same phase (e.g., first post-contrast phase), making automated tumor segmentation challenging. Here, we propose a tumor segmentation method that leverages knowledge of the image acquisition time to modulate model features according to the specific acquisition sequence. We incorporate the acquisition times using feature-wise linear modulation (FiLM) layers, a lightweight method for incorporating temporal information that also allows for capitalizing on the full, variables number of images acquired per imaging study. We trained baseline and different configurations for the time-modulated models with varying backbone architectures on a large public multisite breast DCE-MRI dataset. Evaluation on in-domain images and a public out-of-domain dataset showed that incorporating knowledge of phase acquisition time improved tumor segmentation performance and model generalization. 

\end{abstract}
%

\input{tex/introduction.tex}

\input{tex/method.tex}
\input{tex/results.tex}
\input{tex/discussion.tex}

\input{tex/acknowledgements.tex}

\bibliographystyle{IEEEbib}
\bibliography{references}

\end{document}

%% file: tex/introduction.tex
\section{Introduction}
\label{sec:introduction}


Dynamic contrast-enhanced magnetic resonance imaging (DCE-MRI) of the breast captures the kinetics of contrast agent uptake and washout in different tissues across different time points~\cite{breastmri2019}. Clinically, DCE-MRI is one of the most sensitive modalities for detecting breast cancer and plays an important role not only in screening women at high risk or with dense breasts, but also critical in assessing the extent of tumors ~\cite{aionbreastmri,kuhl1999dynamic}. T1-weighted images are acquired before and multiple times after contrast agent administration to capture a dynamic pattern of enhancement, forming an informative 4D imaging data.


After the injection of contrast agents, different tissues exhibit distinct kinetics. Cancers typically demonstrate fast initial uptake followed by late washout, while benign tissue more often shows a persistent or plateau pattern. These temporal behaviors create contrast and help cancer diagnosis~\cite{breastmri2019}. Radiologists compare the different enhancement levels in pre-contrast and post-contrast images to identify possible lesions. As visualized in~\cref{fig:featuremap}, the tumor region enhances quickly in the first post-contrast image (Fig.~\ref{fig:featuremap}, at t${_1}$). Meanwhile, other tissues' contrast saturates more slowly, but eventually reaches a relatively similar brightness, as in the second post-contrast image (Fig.~\ref{fig:featuremap}, at t${_2}$). Due to individual patient differences and widely varying imaging protocols, including acquisition time of different contrast phases, the appearance of the contrast-enhanced tissues and lesions can largely vary, even within the nominally same imaging phase (e.g., first post-contrast image). This represents a challenge for automated breast tumor segmentation models.
\input{fig/fig_vis}

Previous models for breast tumor segmentation leveraged information of dynamic contrast in different ways. 
Wang et~al.~\cite{wang2021breast} used temporal information through a tumor-sensitive synthesis module to regress post-contrast tumor regions from pre-contrast input. 
Lv et al.~\cite{lv2023diffusion} explicitly modeled the enhancement dynamics through diffusion models to generate images for data augmentation for the segmentation task. 
Yan et al.~\cite{yan2025clinical} used the contrast information by iteratively fusing pre- and post-contrast features across multiple resolutions with dynamic weights.
Zhao et al.~\cite{zhao2024swinhr} used two branches of transformer-based networks as the encoder, one for pre-contrast and one for peak contrast, then fused the information in the decoder.
However, temporal information is often neglected in recent DCE-MRI segmentation works~\cite {li2025hcma,zhang2019deep,giaccaglia2025multiscale}.

Instead, we propose to incorporate temporal information in the DCE-MRI via
Feature-wise Linear Modulation (FiLM)~\cite{perez2018film}. We condition the network on acquisition time by modulating intermediate feature maps with learned parameters. Specifically, each image phase is associated with its corresponding acquisition time, which is encoded by a lightweight conditioning network to produce per-channel scaling and shifting coefficients $(\gamma, \beta)$. These coefficients modulate feature maps in selected layers, allowing the segmentation network to adapt to the temporal dynamics of contrast enhancement. This mechanism enables the model to recognize fast early enhancement typical of malignant lesions while suppressing slowly enhancing benign patterns, without requiring all phases to be explicitly stacked as input channels. Conditioning the representation on continuous acquisition times rather than discrete phase indices, FiLM provides a flexible way to encode enhancement kinetics. It improves robustness to variable numbers of time points and offers a lightweight method to incorporate temporal information.

%% file: fig/fig_vis.tex
\begin{figure}[!t]
    \centering
    \includegraphics[width=1.0\columnwidth]{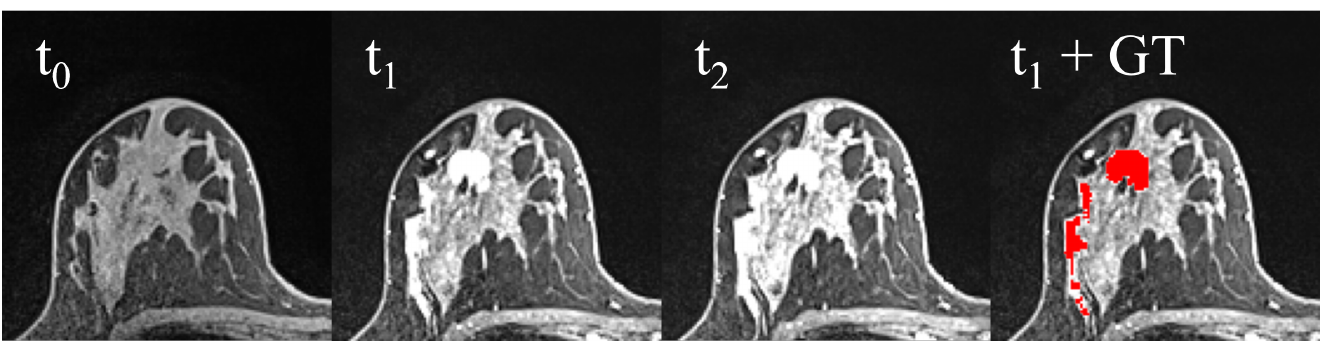}
    \vspace{-2.0\baselineskip}
    \caption{{\bf Visualization of DCE-MRI enhancement.} 
    From left to right are pre-contrast, first post-contrast, second post-contrast, and  expert segmentation of the tumor overlay on the first post-contrast image.} 
    \label{fig:featuremap}
\end{figure}

%% file: tex/method.tex
\section{Methods}
\label{sec:methods}
We propose a method to incorporate the acquisition time of DCE-MRI images to improve breast cancer segmentation. The overall architecture is shown in \cref{fig:method}. Alongside the commonly used decoder-encoder architecture, we generate modulation vectors $\gamma$ and $\beta$ to feature-wise modulate intermediate feature maps in the encoder and decoder blocks, injecting temporal knowledge to the segmentation model. 

\input{fig/fig_method}

\subsection{Time-Informed Feature-Wise Transformations}

Feature-Wise Linear Modulation (FiLM) is a general-purpose conditioning method for deep neural networks~\cite{perez2018film}. We employ FiLM to condition the neural network on the acquisition times of the DCE-MRI sequence. For a feature map $F$ with $C$ channels, a FiLM generator takes in a conditioning vector, in our case, the acquisition times of the corresponding input images $t = [t_1, t_2, t_3]$. The FiLM generator outputs two vectors, $\gamma(t)$ and $\beta(t)$, each with length $C$. We use a neural network with two layers as the FiLM generator, which outputs a vector with dimension $2C$, where the first and second half of the vector correspond to $\gamma(t)$ and $\beta(t)$, respectively.

With the $\gamma(t)$ and $\beta(t)$ vectors, we modulate the feature map in the following way:
$$\operatorname{FiLM}(x) = \gamma(t) \odot x + \beta(t),$$
where $x$ is a feature map with shape $(C \times H\times W \times D)$. For each channel in $C$, we feature-wisely multiply $x$ by the corresponding scalar in $\gamma(t)$ and add by the corresponding scalar in $\beta(t)$ to inject our prior knowledge of acquisition time. The modulated feature map will also have dimension $C \times H\times W \times D$.

\subsection{Using FiLM In Segmentation Architectures}

We use two backbone architectures: \textbf{nnU-Net}~\cite{nnunet} and \textbf{Swin-UNETR}~\cite{swinunetr} as base model. nnU-Net uses a purely convolutional encoder and decoder, and Swin-UNETR uses a hierarchical SwinTransformer encoder and a convolutional decoder. 

We investigate four different strategies of using FiLM in these architectures. As shown in~\cref{fig:method}, we place the FiLM layers (1) after all encoder stages (blue), (2) after all decoder stages (purple), (3) after the bottleneck only (orange), (4) after all encoder stages and all decoder stages (blue + purple + orange). For each of the FiLM layers inserted into the backbone, we have a dedicated FiLM generator, so that the modulation is specific to the feature distribution. 

%% file: fig/fig_method.tex
\begin{figure}[t]
    \centering
    \includegraphics[width=1.0\columnwidth]{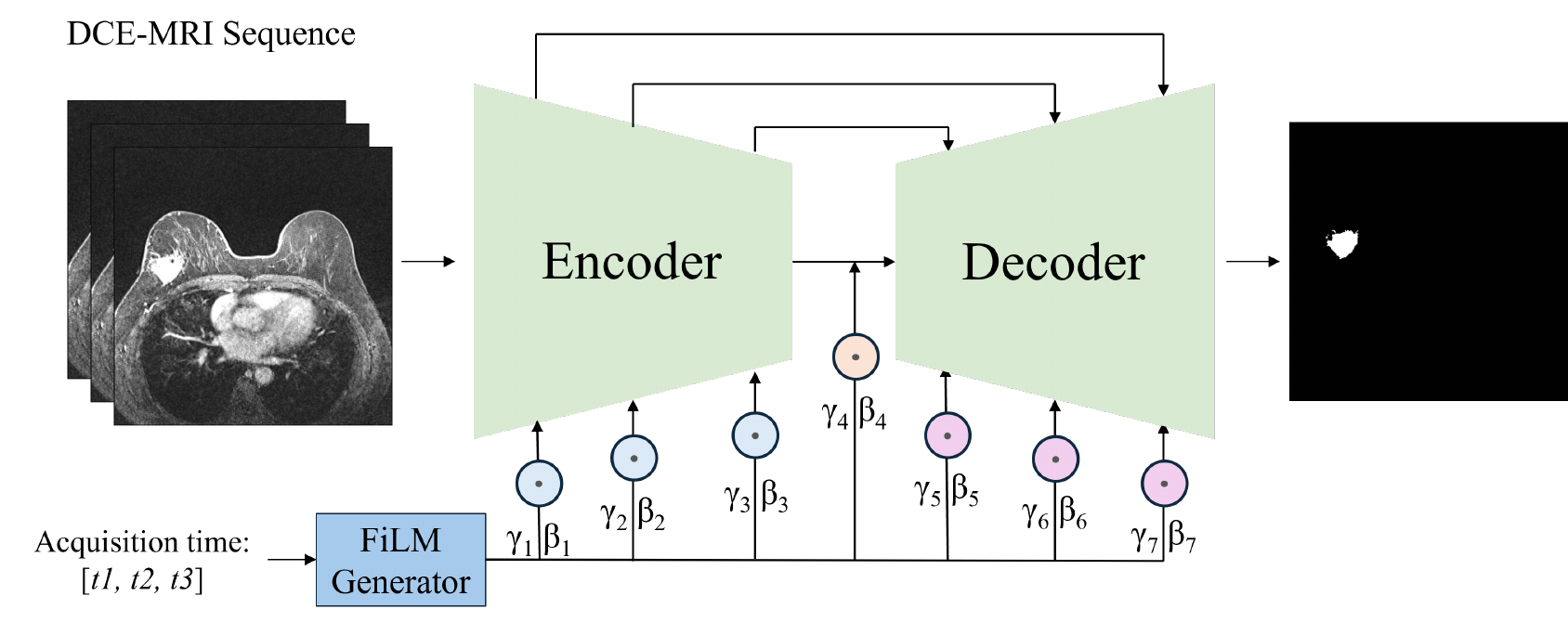}
    \vspace{-1.8\baselineskip}
    \caption{{\bf Proposed Architecture.} 
    A common segmentation architecture often contains an encoder and a decoder block as well as skip connections. We modulate intermediate feature maps in the encoder and decoder stages. We experimented with four different configurations. }
    \label{fig:method}
\end{figure}

%% file: tex/results.tex
\section{Experiments and Results}
\label{sec:results}

\subsection{Experimental Setup}

\paragraph*{Datasets:}
Our primary dataset is the public MAMA-MIA Dataset~\cite{mamamia}. The MAMA-MIA dataset is an accumulated dataset that contains 1506 cases of DCE-MRI images from four breast MRI datasets: ISPY1~\cite{ISPY1}, ISPY2~\cite{ISPY2}, DUKE~\cite{duke_data}, and NACT~\cite{NACT}. In the 1506 cases, we exclude 33 cases where the acquisition time is not available. In the remaining 1473 cases, we randomly choose 200 cases for testing. In the remaining 1273 cases, we randomly split the dataset in 5 folds. In each fold, we use 80\% of the data for training and 20\% for validation. Each case appears exactly once in validation and four times in training across the 5-fold cross-validation. We use the same folds for both nnU-Net-based and Swin-UNETR-based architectures. 

Additionally, we use a public dataset from Yunnan Cancer Hospital~\cite{yunnan_data} for out-of-domain evaluation. The dataset contains 100 cases with DCE-MRI sequence and expert annotations, with acquisition times of ``1-2 minutes" per phase. We take the average of this range and set the pre-contrast phase as 0 seconds, the first post-contrast phase as 90 seconds, the second-post contrast phase as 180 seconds, etc. 


\paragraph*{Data Pre-Processing:}
We \begin{inparaenum}[(1)]
    \item applied N-4 bias field correction ~\cite{n4itk} to all the images;  
    \item resampled all the images to $1.0 \times1.0\times1.0 \text{mm}^3$ using B-spline interpolation;
    \item normalized image intensities per-study using the minimum and 99 percentile intensity for each subject. 
\end{inparaenum}

\paragraph*{Training:}
Cases in the dataset contain varying numbers of DCE-MRI phases, with a minimum of three (pre-contrast, first post-contrast, and second post-contrast). To maintain a consistent input dimensionality across all models, each training sample is constructed using three channels. The first two channels are always assigned to the pre-contrast and first post-contrast phases, as these typically provide the most informative enhancement characteristics and were used to annotate the tumors. The third channel is selected from the remaining later post-contrast phases.

For cases with more than three available phases, multiple samples are generated by pairing the fixed pre-contrast and first post-contrast phases with each additional phase. For example, if a case contains a pre-contrast and four post contrast phases, three samples are created: [pre, first, second], [pre, first, third], and [pre, first, fourth]. The corresponding acquisition time associated with each selected phase is included as a conditioning vector. This procedure ensures consistent network input while leveraging the full temporal information available for each case.

\paragraph*{Implementation Details:}
For nnU-Net, we use the official implementation~\cite{nnunet}. We choose the 3D full-resolution configuration and use their automatic preprocessing pipeline. The images are sliced into $3 \times 128 \times128\times128$ voxel patches. We train with the stochastic gradient descent optimizer. The models are trained for 1000 epochs.
For Swin-UNETR~\cite{swinunetr}, we use the MONAI framework for implementation. The images are sliced into $128 \times128\times128$ voxel patches. We train with the AdamW optimizer~\cite{adamw}. The models are trained for 100 epochs. For both architectures, we used dice and cross-entropy loss. 

\paragraph*{Evaluation Metrics:}

We evaluate the effectiveness of acquisition time modulation for each dataset using the Dice score and the $95^{\text{th}}$ percentile Hausdorff distance (HD95). We report both the average Dice score (Dice) and the $10^{\text{th}}$ percentile Dice score (Dice10) across test samples. While the mean Dice summarizes overall performance, Dice10 captures performance on the most challenging cases by indicating the Dice value below which only 10\% of test cases fall (i.e., in 90\% of cases the Dice is at least Dice10). This provides a robust characterization of “tail” behavior that is less sensitive to single outliers than the minimum Dice and is more informative about model reliability on difficult or atypical scans than reporting the mean alone.
We use two-tailed paired t-tests ($\alpha$=$0.05$) to assess for significant differences between models.


\input{fig/table_rui}
\subsection{Results}
\paragraph*{In-Domain Results on MAMA-MIA Dataset:}
\cref{tab:results_table} reports segmentation performance on the MAMA-MIA dataset for all FiLM configurations and both backbone architectures. For nnU-Net, incorporating acquisition-time conditioning consistently improves performance over the baseline. Modulating all encoder and decoder stages yields  the highest Dice score (0.774), the best tail-case accuracy as reflected by Dice10 (0.473), and the lowest HD95 (35.0 mm). For Swin-UNETR, FiLM similarly improves performance, with the best results obtained when applying modulation at all stages (Dice 0.759, Dice10 0.487, HD95 26.7 mm). Paired two-tailed t-tests confirm that using FiLM is statistically different from the no-modulation baseline. These results show that conditioning the feature representation on continuous acquisition times enhances segmentation accuracy.

\paragraph* {Generalization to Yunnan Dataset:}
We further evaluate the generazability of all models on the external Yunnan dataset. As shown in Table 1, incorporating acquisition-time modulation improves the performance of both architectures, despite the approximated temporal annotations in this dataset. For nnU-Net, modulating all encoder and decoder stages again yields the strongest performance, with the highest Dice (0.762), best Dice10 (0.491), and lowest HD95 (74.3 mm). While nnU-Net maintains relatively strong performance on the Yunnan dataset, Swin-UNETR shows substantial degradation. One possible explanation for limited robustness to out-of-distribution data is due to the transformer-based encoder, which is requires large amounts of data for effective learning. Thus, the transformer-based architecture is more sensitive to changes in data distribution with a weaker inductive bias stemming from its architecture compared to the convolution-based architecture.

\paragraph* {Effect of FiLM Placement:}
Across both backbones, the choice of where FiLM layers are inserted influences performance. For nnU-Net, modulating all encoder and decoder stages consistently yields better results. This suggests that temporal information from acquisition times is most useful when injected throughout the full hierarchy. Encoder-only and decoder-only modulation provide moderate improvements, while bottleneck-only modulation offers limited improvement. For Swin-UNETR, encoder-only modulation performs best on the external dataset, and all-layer modulation slightly improves performance on MAMA-MIA. This could be caused by transformer encoders relying heavily on long-range token interactions; conditioning early on feature embeddings appears more effective than modulating later decoding stages.

\input{fig/fig_qualitative}

\paragraph* {Qualitative Evaluation:}
Qualitative example results from the Yunnan dataset in Fig.\ 3 illustrate how acquisition-time modulation affects tumor delineation across different FiLM placements. The baseline model tends to over-segment and exhibits fragmented predictions, while FiLM-equipped models produce more coherent lesions with cleaner boundaries. Encoder and decoder modulation each reduce spurious false positives, and the combined encoder–decoder configuration yields the closest match to the ground truth. Bottleneck-only modulation offers modest improvements but lacks the spatial refinement observed when FiLM is applied across multiple stages.

\paragraph* {Overall Findings:}
Across datasets and architectures, these examples reflect the broader quantitative trends: temporal conditioning leads to more stable representations of enhancement kinetics and improves segmentation quality, with the strongest gains seen when modulation is applied throughout the network hierarchy. Taken together, these results show that acquisition-time FiLM provides a lightweight yet effective mechanism to enhance both segmentation accuracy and robustness in DCE-MRI tumor analysis.

%% file: fig/table_rui.tex
\begin{table*}[t]
\centering
\small
\begin{tabular}{ll*{6}{c}}
\toprule
& & \multicolumn{3}{c}{MAMA-MIA} & \multicolumn{3}{c}{Yunnan} \\
\cmidrule(lr){3-5}\cmidrule(lr){6-8}
Backbone & Time Modulation & Dice($\uparrow$) & Dice10($\uparrow$) & HD95($\downarrow$) & Dice($\uparrow$) & Dice10($\uparrow$) & HD95($\downarrow$) \\
\midrule
\multirow{5}{*}{nnU-Net}
& None (Baseline) & 0.766 $\pm$ 0.203 & 0.439 & ~~37.0 $\pm$ 55.7 & 0.736 $\pm$ 0.188 & 0.447 & 90.4 $\pm$ 62.0\\
& Encoder         & 0.767 $\pm$ 0.201 & 0.466 & ~~38.2 $\pm$ 57.3 & ~~0.749 $\pm$ 0.171* & 0.468 & ~~88.3 $\pm$ 60.4* \\
& Decoder         & 0.768 $\pm$ 0.200 & 0.468 & ~~38.5 $\pm$ 57.8 & ~~0.741 $\pm$ 0.181* & 0.463 & 89.5 $\pm$ 60.5\\
& Bottleneck      & 0.765 $\pm$ 0.202 & 0.448 & ~~40.3 $\pm$ 58.8 & ~~0.757 $\pm$ 0.169* & 0.480 & ~~79.1 $\pm$ 60.8* \\
& All             & ~~\textbf{0.774 $\pm$ 0.200*} & \textbf{0.473} & ~~\textbf{35.0 $\pm$ 53.6*} & ~~\textbf{0.762 $\pm$ 0.165*} & \textbf{0.491} & ~~\textbf{74.3 $\pm$ 61.9*} \\
\midrule
\multirow{5}{*}{SwinUNETR}
& None (Baseline) & 0.754 $\pm$ 0.220 & 0.380 & 28.3 $\pm$ 39.9 & 0.375 $\pm$ 0.233 & 0.104 & 126.2 $\pm$ 40.7\\
& Encoder         & 0.757 $\pm$ 0.210 & 0.450 & 27.6 $\pm$ 40.0 & ~~\textbf{0.418 $\pm$ 0.227*} & 0.114 & ~~\textbf{122.5 $\pm$ 41.5*} \\
& Decoder         & 0.755 $\pm$ 0.215 & 0.393 & 30.1 $\pm$ 43.6 & ~~0.385 $\pm$ 0.226* & 0.102 & ~~124.6 $\pm$ 41.4* \\
& Bottleneck      & 0.754 $\pm$ 0.208 & 0.441 & 29.5 $\pm$ 41.7 & ~~0.404 $\pm$ 0.222* & \textbf{0.143} & ~~124.3 $\pm$ 42.3* \\
& All             & ~~\textbf{0.759 $\pm$ 0.206* } & \textbf{0.487} & ~~\textbf{26.7 $\pm$ 37.6*} & ~~0.394 $\pm$ 0.234* & 0.102 & 126.4 $\pm$ 40.4 \\
\bottomrule
\end{tabular}
\vspace{-4mm}
\caption{Tumor segmentation results across datasets. HD95 = Hausdorff distance, Dice = Average Dice score, and Dice10 = the $10^{\text{th}}$ percentile Dice score. Best result for each dataset is in bold. $^*$ denotes significant difference from the baseline model with no time modulation  ($p < 0.05$).}
\label{tab:results_table}
\vspace{-4mm}
\end{table*}

%% file: fig/fig_qualitative.tex
\begin{figure}[!t]
    \centering
    \includegraphics[width=1.0\columnwidth]{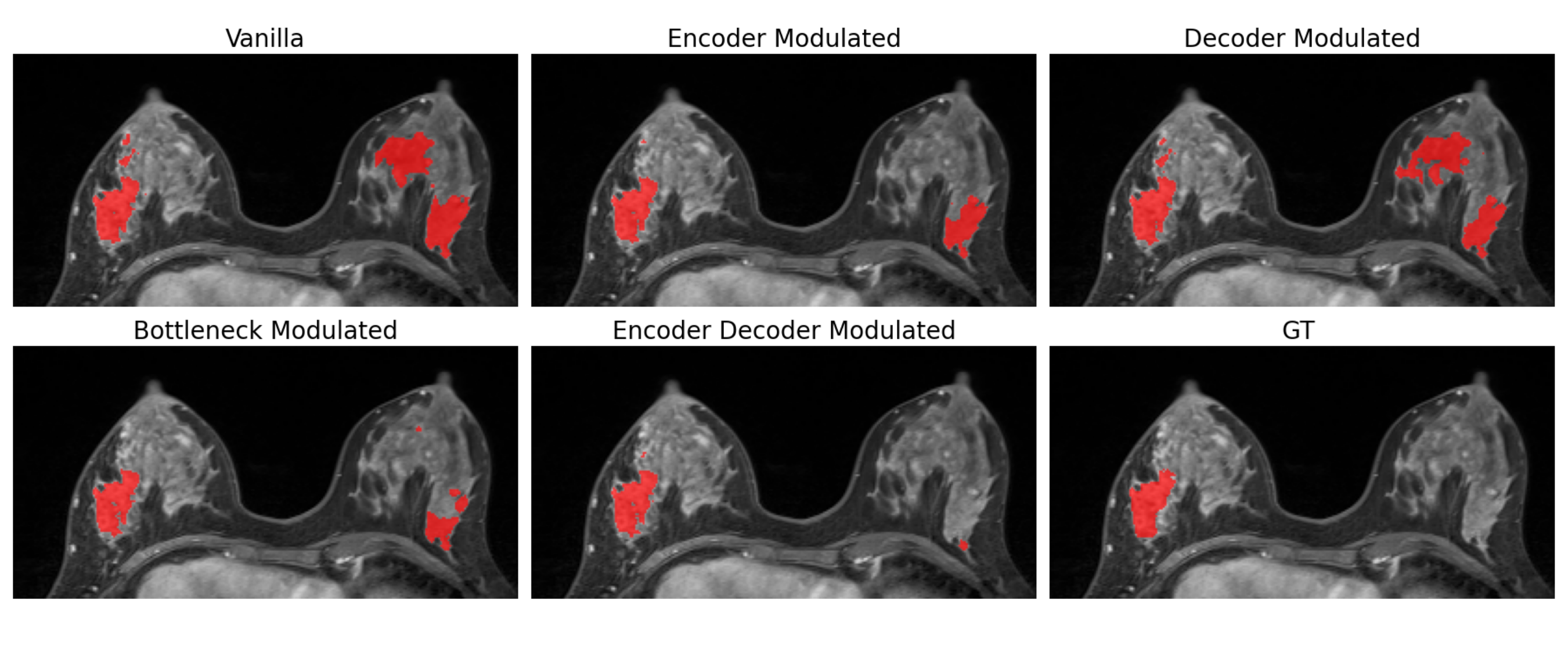}
    \vspace{-2.0\baselineskip}
    \caption{{\bf Qualitative comparison of tumor segmentation across FiLM configurations.} 
DCE-MRI slice showing predictions from the baseline nnU-Net (Vanilla) and four FiLM placement strategies (Encoder, Decoder, Bottleneck, and Encoder–Decoder modulation). Baseline model yields most false positives without infused temporal enrichment information. The best-performing configuration (Encoder–Decoder Modulated) yields much fewer false positives. Ground truth (GT) is shown in the bottom right.} 
    \label{fig:segexample}
    \vspace{-1.0\baselineskip}
\end{figure}

%% file: tex/discussion.tex
\section{Discussion and Conclusion}
\label{sec:discussion}
By conditioning segmentation networks on continuous acquisition times, our approach leverages the intrinsic temporal dynamics of DCE-MRI. FiLM allows the model to learn feature representations beyond static intensity patterns. Unlike conventional architectures that treat each phase as an independent input channel, FiLM directly modulates intermediate features, enabling the network to encode enhancement kinetics relevant to malignancy. Our results demonstrate that acquisition-time modulation improves segmentation accuracy. In particular, gains in Dice10 indicate that FiLM reduces the frequency of severe failure cases and strengthens robustness on the most challenging scans.
Both nnU-Net and Swin-UNETR benefit from temporal modulation, showing improved performance. This suggests that FiLM conditioning offers a lightweight means of incorporating temporal information for the segmentation task. The method also shows improvements when applied with estimated acquisition timing.
FiLM layers add few parameters yet provide stable time-aware representations that improve segmentation results in both the overlap-based metric and distance-based metric. Future work will further explore conditioning strategies and integration with kinetic modeling to strengthen temporal reasoning.
Overall, acquisition-time conditioning provides a practical mechanism to exploit DCE-MRI dynamics, improving segmentation accuracy with minimal architectural changes.

%% file: tex/acknowledgements.tex
\paragraph*{Acknowledgements}
This work was supported by NIH grant R21 EB032950 and ARPA-H grant D24AC00156.


\paragraph*{Compliance with Ethical Standards}
This research study was conducted retrospectively in part using human subject data made available in open access: the MAMA-MIA dataset by Garrucho et al.\cite{mamamia} and the Yunnan dataset by Zhang et al.\cite{yunnan_data}. Ethical approval was not required as confirmed by the license attached with the open access data. 
